\documentclass[10pt,twocolumn]{article}

\usepackage[margin=1in]{geometry}
\usepackage{times}
\usepackage{graphicx}
\usepackage{booktabs}
\usepackage{amsmath,amssymb}
\usepackage[numbers]{natbib}   
\usepackage[hidelinks]{hyperref}

\graphicspath{{./}{./images/}}

\title{Relational Graph Modeling for Credit Default Prediction: Heterogeneous GNNs and Hybrid Ensemble Learning}
\author{
  Yvonne Yang \\
  University of Illinois Urbana-Champaign \\
  \texttt{yuweny4@illinois.edu}
  \and
  Eranki Vasistha \\
  University of Illinois Urbana-Champaign \\
  \texttt{veranki2@illinois.edu}
}

\date{} 

\begin{document}
\maketitle

\begin{abstract}
Credit default risk arises from complex interactions among borrowers, financial institutions, and transaction-level behaviors. While strong tabular models remain highly competitive in credit scoring, they may fail to explicitly capture cross-entity dependencies embedded in multi-table financial histories. In this work, we construct a massive-scale heterogeneous graph containing over 31 million nodes and more than 50 million edges, integrating borrower attributes with granular transaction-level entities such as installment payments, POS cash balances, and credit card histories.

We evaluate heterogeneous graph neural networks (GNNs), including heterogeneous GraphSAGE and a relation-aware attentive heterogeneous GNN, against strong tabular baselines. We find that standalone GNNs provide limited lift over a competitive gradient-boosted tree baseline, while a hybrid ensemble that augments tabular features with GNN-derived customer embeddings achieves the best overall performance, improving both ROC-AUC and PR-AUC. We further observe that contrastive pretraining can improve optimization stability but yields limited downstream gains under generic graph augmentations. Finally, we conduct structured explainability and fairness analyses to characterize how relational signals affect subgroup behavior and screening-oriented outcomes.
\end{abstract}

\begin{abstract}
Credit default risk arises from complex interactions among borrowers, financial institutions, and transaction-level behaviors. While strong tabular models remain highly competitive in credit scoring, they may fail to explicitly capture cross-entity dependencies embedded in multi-table financial histories. In this work, we construct a massive-scale heterogeneous graph containing over 31 million nodes and more than 50 million edges, integrating borrower attributes with granular transaction-level entities such as installment payments, POS cash balances, and credit card histories.

We evaluate heterogeneous graph neural networks (GNNs), including heterogeneous GraphSAGE and a relation-aware attentive heterogeneous GNN, against strong tabular baselines. We find that standalone GNNs provide limited lift over a competitive gradient-boosted tree baseline, while a hybrid ensemble that augments tabular features with GNN-derived customer embeddings achieves the best overall performance, improving both ROC-AUC and PR-AUC. We further observe that contrastive pretraining can improve optimization stability but yields limited downstream gains under generic graph augmentations. Finally, we conduct structured explainability and fairness analyses to characterize how relational signals affect subgroup behavior and screening-oriented outcomes.
\end{abstract}


\maketitle

\section{Introduction}
Credit default prediction remains a core problem for lenders because even small gains in ranking accuracy can translate into fewer charge-offs and broader, fairer access to credit. We study borrower-level default risk using the Home Credit Default Risk (HCDR) dataset, which joins multi-table credit histories (application, bureau, credit-card, POS/Cash, and installments) with behavioral attributes. These data are high-dimensional, heterogeneous, and implicitly relational, motivating approaches that balance predictive accuracy, calibration, and operational interpretability.

Traditional credit scoring relies on transparent linear models, while modern tabular learners—especially gradient-boosted decision trees—frequently provide state-of-the-art performance on imbalanced financial datasets \cite{p1,p2}. Handling severe class imbalance is essential for risk ranking at low approval rates; comparative studies emphasize stratified evaluation and cost-sensitive learning, with oversampling or boosting-based strategies often outperforming naive undersampling as imbalance grows \cite{p3,p4}. Deep learning has also been explored for credit scoring, leveraging convolutional, recurrent, or attention mechanisms to capture complex behavioral signals, though results can be dataset-specific and sensitive to architecture and validation design \cite{p5,p6}.

Beyond instance-only features, many financial risks are relational: defaults may co-vary across shared employers, addresses, devices, merchants, or co-borrowers. Graph Neural Networks (GNNs) are designed to exploit such dependencies, and surveys document their growing use in credit and broader financial risk modeling, where graphs are constructed from interaction logs, transaction networks, or entity co-occurrences \cite{gnninfinance,p7,p8}. 

From a practical perspective, financial institutions already rely on highly optimized tabular pipelines. Understanding when relational modeling provides complementary value—rather than assuming universal gains—is therefore critical for real-world adoption.

In this project, we first establish a strong, auditable baseline on HCDR using regularized logistic regression and gradient-boosted trees under stratified 5-fold cross-validation with early stopping. Our evaluation emphasizes ranking quality under class imbalance (ROC-AUC, PR-AUC), with additional diagnostics such as KS statistics, operational cutoffs, and calibration analyses used during model development and validation. After establishing this baseline, we implement a GNN that encodes cross-table relations among applicants, accounts, and bureau records to test whether relational inductive bias yields measurable lift over the baseline under identical preprocessing and metrics. This setup lets us quantify the marginal value of relational modeling under realistic constraints aligned with industry practice.

\section{Related Work}

\subsection{Quantitative Credit Risk Modeling}
Credit risk modeling has a long tradition in both consumer and corporate finance. Classical credit scoring methods, including logistic regression, support vector machines, and k-nearest neighbors, have long served as transparent baselines in regulated risk prediction settings \cite{p1,p2}. Comparative studies across multiple datasets report that no single classical model dominates across all evaluation criteria—such as accuracy, sensitivity, specificity, or AUC—highlighting the importance of dataset-aware model selection and multi-metric evaluation \cite{p1,p2}.

On modern, highly imbalanced tabular credit datasets, gradient-boosted decision tree models (e.g., XGBoost and LightGBM) consistently outperform linear baselines and generic multilayer perceptrons in ranking-based metrics, establishing them as strong defaults for production credit scoring pipelines \cite{p1,p2}. Beyond ROC-AUC, the literature emphasizes PR-AUC and Kolmogorov–Smirnov statistics, as well as operational measures such as recall at fixed approval rates, to better reflect real-world risk screening objectives \cite{p1,p7}. Calibration quality is increasingly highlighted to ensure that ranking performance translates into reliable probability estimates for downstream decision-making and portfolio management \cite{p7,p8}.

\subsection{Financial Networks and Systemic Risk}
A complementary body of work studies financial risk through the lens of economic and financial networks, where interdependencies among agents can amplify shocks and generate systemic vulnerabilities. Seminal research models financial systems as networks of interconnected institutions, demonstrating how distress can propagate through balance-sheet linkages and lead to non-linear, system-wide failures \cite{Schweitzer2009,Battiston2012}. Network topology, node centrality, and exposure heterogeneity are shown to play a critical role in determining systemic importance and tail risk.

While this literature primarily focuses on macro-level systemic stability, it underscores a fundamental insight: financial risk is inherently relational. Our work builds on this perspective by modeling borrower-level credit risk within a heterogeneous financial network, where customers, prior loans, and credit bureau records form a structured multi-entity system. This micro-level view complements macro-network approaches and connects quantitative credit modeling with network-based risk analysis.

\subsection{Machine Learning for Credit Risk at Scale}
For large, noisy credit datasets such as Home Credit, domain-informed feature engineering remains a critical determinant of model performance. Prior work reports that capacity and affordability ratios, delinquency indicators, repayment regularity metrics, and revolving credit utilization features substantially improve ranking quality for both boosted trees and calibrated linear models \cite{p5,p6}. Ensemble-based importance rankings provide practical feature selection and interpretability compared to purely unsupervised dimensionality reduction, often yielding compact, high-gain feature subsets \cite{p5,p6}.

Severe class imbalance is a defining characteristic of default prediction. Synthetic oversampling techniques such as SMOTE and its variants mitigate bias toward majority classes by generating minority samples along nearest-neighbor manifolds \cite{p3}. Hybrid approaches—including Borderline-SMOTE, SMOTE–Tomek, and cluster-aware resampling—further refine decision boundaries and reduce noise, particularly for tree-based learners under skewed class distributions \cite{p3,p4,p5}. Cost-sensitive learning and stratified validation remain complementary best practices regardless of the resampling strategy \cite{p3,p4}.

\subsection{Graph Neural Networks for Financial Risk and Fraud}
Many forms of financial risk are relational: defaults may co-vary through shared employers, addresses, financial products, devices, or behavioral patterns. Recent surveys and empirical studies document growing use of graph neural networks (GNNs), including GCN, GraphSAGE, and GAT variants, for fraud detection and credit risk modeling \cite{gnninfinance,p7,p8}. When relational signal is informative, GNNs are shown to improve ranking quality over strong tabular baselines by capturing higher-order dependencies that are difficult to encode in independent feature representations \cite{gnninfinance,p7}.

However, existing studies often emphasize graph model accuracy in isolation or rely on homogeneous graph assumptions, limiting insight into their practical value in production risk systems. In contrast, our work focuses on systematic comparison against optimized tabular baselines and examines explainability and fairness implications of relational modeling—an increasingly important consideration in financial decision-making contexts.

\subsection{Fairness and Explainability in Credit Decision Systems}
As machine learning models are increasingly deployed in high-stakes financial applications, concerns around fairness, transparency, and accountability have gained prominence. Prior work documents how algorithmic credit systems may exhibit disparate impacts across demographic groups and emphasizes the need for explainable models that support regulatory oversight \cite{Barocas2016,Fuster2022}. Recent studies highlight that representation learning choices can influence both predictive performance and subgroup behavior, underscoring the importance of auditing model components beyond aggregate metrics.

\section{Dataset and Exploratory Data Analysis}

\subsection{Relational Dataset Overview}
We use the Home Credit Default Risk (HCDR) dataset, a multi-table borrower-level panel commonly used in credit risk modeling. The primary training table \texttt{application\_train} contains \textbf{307,511 rows and 122 columns}, capturing applicant demographics, financial capacity, and credit history. The dataset follows a canonical relational schema linked by specific keys:
\begin{itemize}
    \item \texttt{SK\_ID\_CURR}: The primary key joining application rows to all borrower-level aggregates.
    \item \texttt{SK\_ID\_BUREAU}: Links the \texttt{bureau} table to \texttt{bureau\_balance} history.
    \item \texttt{SK\_ID\_PREV}: Links \texttt{previous\_application} to its monthly child tables (\texttt{POS\_CASH\_balance}, \texttt{installments\_payments}, \texttt{credit\_card\_balance}).
\end{itemize}

This relational structure reflects a realistic financial data warehouse design, where borrower-level outcomes are influenced by historical applications and granular transactional behavior.

\subsection{Graph Construction and Scale}

We transform the relational schema into a heterogeneous entity--relation graph to explicitly model dependencies across borrowers, credit instruments, and transactional behaviors. The resulting graph consists of six node types: customers, previous applications, bureau records, installment payments, credit card balances, and POS cash balances. Edges encode semantic financial relations such as application history, external credit exposure, and repayment dynamics.

When fully materialized, this heterogeneous graph contains approximately \textbf{31 million nodes} and over \textbf{50 million edges} (including bidirectional relations). While this scale precludes full-batch training, it enables the model to capture fine-grained behavioral signals that are not accessible through tabular aggregation alone.

To ensure computational feasibility, we adopt mini-batch neighborhood sampling during training, constructing localized subgraphs centered on customer nodes. This approach preserves multi-hop relational risk signals while maintaining tractable memory usage and runtime performance.

\subsection{Exploratory Data Analysis (EDA)}
Before constructing graph-based models, we conduct exploratory data analysis to examine feature distributions, class imbalance, and missingness patterns in the primary application table.

\begin{figure}[hbt!]
  \centering
  \includegraphics[width=0.9\linewidth]{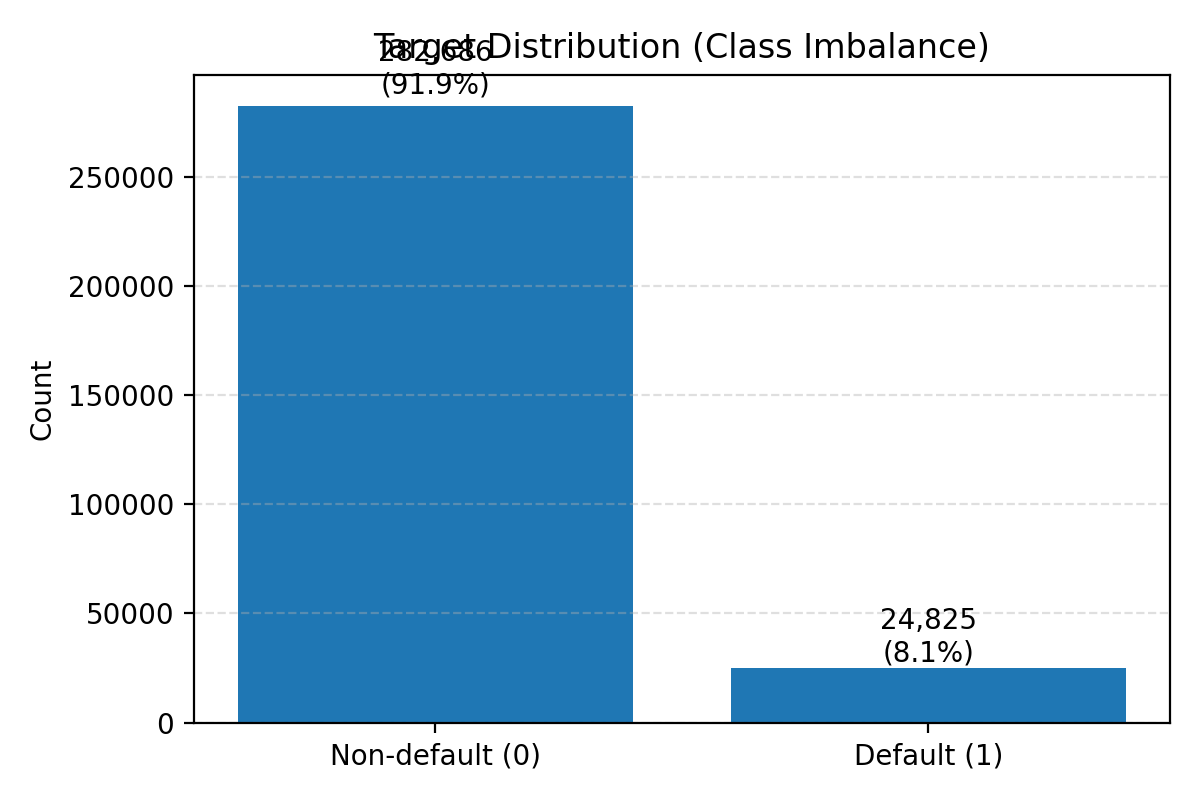}
  \caption{Target distribution (\texttt{TARGET}) in \texttt{application\_train}, highlighting significant class imbalance.}
  \label{fig:target-dist}
\end{figure}

\paragraph{Target Distribution and Imbalance}
The task is binary default prediction using the \texttt{TARGET} variable in \texttt{application\_train}. As shown in Figure \ref{fig:target-dist}, the class balance is highly skewed ($\sim$8\% positive class). Consequently, downstream evaluation focuses on rank-oriented metrics (AUC) rather than accuracy.

\begin{figure}[hbt!]
  \centering
  \includegraphics[width=0.9\linewidth]{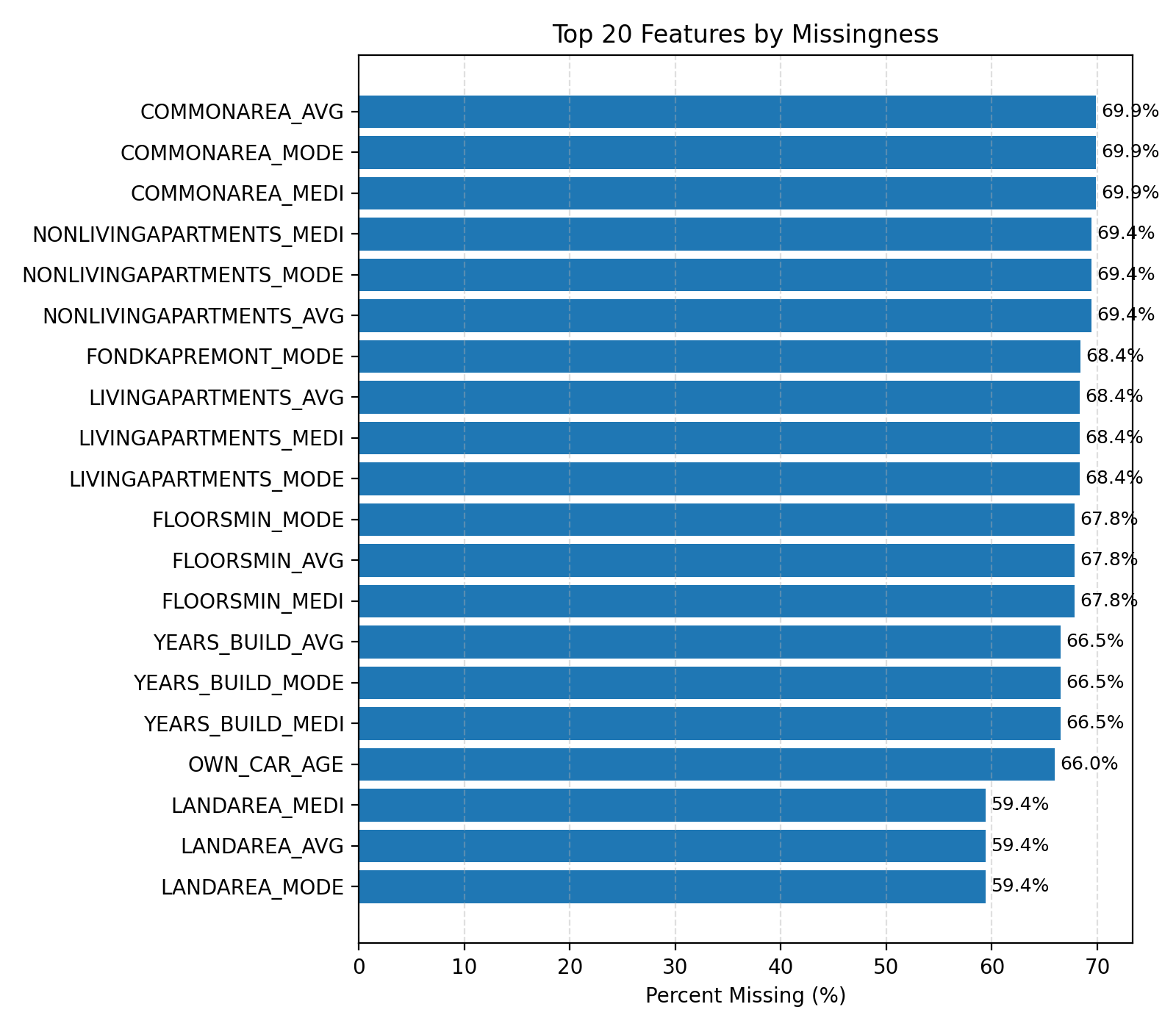}
  \caption{Top-20 features by missingness in the merged frame. 69 columns with $>80\%$ missingness are dropped.}
  \label{fig:missing-top20}
\end{figure}

\paragraph{Missingness Profile}
We analyzed column-wise missing rates across the merged feature set. As illustrated in Figure \ref{fig:missing-top20}, several features exhibit extreme sparsity. Columns with missingness $>$\;80\% were dropped, resulting in the removal of \textbf{69} high-missing columns. This pruning step precedes scaling and dimensionality reduction.

\paragraph{Skewness and Outliers}
We measured univariate skewness and flagged features with $|\text{skew}| > 5$, identifying \textbf{202} highly skewed features (mostly financial magnitudes). Additionally, we computed IQR-based outlier fractions (using the $1.5 \times \text{IQR}$ rule) to identify heavy-tailed distributions typical of income and credit amount variables.

\paragraph{Scaling and Transforms}
To handle this skewness, we implemented a robust preprocessing pipeline:
\begin{itemize}
  \item \textbf{Log Transformation:} A $\log(1+x)$ transform is applied to heavy-tailed monetary and ratio features (columns containing \texttt{AMT}, \texttt{SUM}, \texttt{RATIO}).
  \item \textbf{Robust Scaling:} These transformed features are then scaled using \texttt{RobustScaler} to mitigate the influence of outliers.
  \item \textbf{Standard Scaling:} All remaining numeric features undergo simple standardization (\texttt{StandardScaler}).
\end{itemize}
Diagnostic checks confirm that this mixed strategy yields near-standardized statistics (mean $\approx$ 0.10, std $\approx$ 1.05) suitable for neural networks.

\paragraph{PCA Diagnostics}
After median imputation, we fit Principal Component Analysis (PCA) on the standardized matrix. We found that retaining 95\% of the variance requires \textbf{111} principal components. This result quantifies the redundancy in the feature space and provides a target dimensionality for compact models.

\paragraph{Preprocessing Summary}
Guided by these diagnostics, the final baseline pipeline consists of: (i) dropping high-missing columns, (ii) median imputation, (iii) targeted log-robust scaling, and (iv) optional PCA. The resulting merged, borrower-level frame (\texttt{merged file}) serves as the input for our baseline models and the reference feature space for the GNN ensemble.


\section{Methodology and Implementation}

\subsection{Tabular Pipeline (Feature Engineering)}
These engineered features are used for the tabular baselines and for the hybrid ensemble; graph models are trained on the heterogeneous graph described in Section 3.

Starting from \texttt{application\_train} as the labeled table, we join borrower histories from the monthly panels and prior-credit records to build a single, analysis-ready matrix at the \texttt{SK\_ID\_CURR} level. Specifically, we aggregate \texttt{installments\_payments}, \texttt{credit\_card\_balance}, and \texttt{POS\_CASH\_balance} by \texttt{SK\_ID\_PREV} and then roll them up to the borrower via \texttt{SK\_ID\_CURR}; identifiers are used strictly for joins and dropped before modeling. 

The engineered features follow two complementary themes: repayment behavior and affordability/capacity.
\begin{itemize}
    \item \textbf{Repayment Behavior:} We compute the per-installment \textit{payment ratio} ($\text{AMT\_PAYMENT} / \text{AMT\_INSTALMENT}$) and a lateness indicator comparing due dates to actual payment dates. These are summarized over time as means, standard deviations, and late-rate aggregates to capture both regularity and slippage in servicing debt.
    \item \textbf{Revolving Behavior:} For credit cards, we form a utilization proxy ($\text{AMT\_BALANCE} / \text{AMT\_CREDIT\_LIMIT\_ACTUAL}$) and summarize its level and variability.
    \item \textbf{Affordability:} At the applicant level, we encode affordability through ratios such as:
    \begin{equation}
        \text{CREDIT\_INCOME} = \frac{\text{AMT\_CREDIT}}{\text{AMT\_INCOME\_TOTAL}}
    \end{equation}
    \begin{equation}
        \text{ANNUITY\_INCOME} = \frac{\text{AMT\_ANNUITY}}{\text{AMT\_INCOME\_TOTAL}}
    \end{equation}
    \begin{equation}
        \text{LOAN\_PER\_FAM} = \frac{\text{AMT\_CREDIT}}{\text{CNT\_FAM\_MEMBERS}}
    \end{equation}
\end{itemize}

Where both burden and delinquency signals are available, we also combine them into a simple ``payment burden'' indicator by multiplying the mean payment ratio with the borrower’s late rate. A light target-encoding pass adds one numeric feature per selected categorical field (e.g., education, occupation), mapping each category to its in-sample default rate; these encodings are appended with a \texttt{\_TE} suffix to keep lineage explicit.

\subsection{Tabular Baselines and Training Protocol}
We ground the study on two complementary baselines that are widely used in credit scoring: gradient-boosted decision trees via LightGBM and a regularized logistic regression. Both are trained on the same borrower-level feature matrix to ensure a fair comparison, and both use the identical stratified five-fold protocol (\texttt{random\_state}=42). To prevent information leakage, all preprocessing that depends on the training data statistics (e.g., imputations or scalers for the linear model) is fitted within each fold and applied to the corresponding validation split; out-of-fold predictions are then retained for later analysis.

\paragraph{LightGBM (Gradient-Boosted Decision Trees)}
We adopt LightGBM because heterogeneous, high-cardinality tabular features with missing values and non-linear effects are common in credit data, and boosted trees tend to model such structure with minimal manual interaction design. Our configuration balances model capacity and generalization: gradient boosting with a conservative learning rate (0.02) and \texttt{num\_leaves}=34 encourages gradual improvements while keeping individual trees from becoming overly complex; \texttt{min\_data\_in\_leaf}=100 combats leaf-level overfitting on rare patterns. Subsampling is applied on both rows and columns (\texttt{bagging fraction} and \texttt{feature\_fraction} at 0.85) to further reduce variance, and we regularize with $\lambda_1=1.0$ and $\lambda_2=2.0$. We allow up to 10{,}000 boosting iterations but engage early stopping after 200 rounds without AUC improvement on the in-fold validation set, which effectively tunes the number of trees to the data. Class imbalance is handled by setting \texttt{scale\_pos\_weight} to the fold-specific ratio of negatives to positives. A practical advantage in this pipeline is that LightGBM consumes \texttt{NaN} natively and learns split directions that route missing values consistently, avoiding heavy imputations.

\paragraph{Logistic Regression (with Scaling and Calibration)}
Logistic regression serves as a transparent, well-calibrated linear baseline that is easy to audit and deploy. Because linear models are sensitive to feature scale and multicollinearity, we standardize the numeric columns within each training fold using \texttt{StandardScaler} (mean-centering disabled to preserve sparsity) and rely on the correlation pruning performed during feature engineering. Heavy-tailed financial magnitudes are paired with their \texttt{log1p} counterparts so the model can approximate simple elasticities. We fit the \texttt{lbfgs} solver with $C=1.0$ and \texttt{max\_iter}=1000, which provides a reasonably strong $L_2$-regularized baseline. Missing numeric values are filled with simple, fold-fitted statistics (zero or median) before scaling. We also produce an isotonic-calibrated variant via \texttt{CalibratedClassifierCV}, trained strictly on held-out folds to ensure calibration does not leak information.

To ensure fair comparison across modeling paradigms, we use the same train/validation/test split for all models. For graph-based models, splits are applied at the customer-node level using the same random seed, while all relational context is preserved during message passing. This protocol ensures that performance differences reflect modeling choices rather than data partitioning effects.

\subsection{Heterogeneous Graph Construction}
To capture non-linear structural relationships in credit history data, we construct a massive-scale heterogeneous graph from the relational tables described in Section~3. The resulting graph contains approximately \textbf{31 million nodes} and over \textbf{50 million edges}, enabling the model to learn from granular transaction-level behaviors.

\begin{itemize}
\item \textbf{Node Types ($\mathcal{V}$):} We identify six entity types to preserve financial activity granularity:
\begin{enumerate}
\item \textbf{Customer:} Current loan applicants (\texttt{SK\_ID\_CURR}).
\item \textbf{Bureau Record:} External credit history (\texttt{SK\_ID\_BUREAU}).
\item \textbf{Previous Application:} Historical internal loans (\texttt{SK\_ID\_PREV}).
\item \textbf{Installment Payment:} Granular repayment events.
\item \textbf{POS Cash Balance:} Monthly POS or cash loan snapshots.
\item \textbf{Credit Card Balance:} Revolving credit usage histories.
\end{enumerate}
\item \textbf{Edge Types ($\mathcal{E}$):} Directed edges follow foreign-key relations (e.g., \textit{Previous Application} $\rightarrow$ \textit{Installment}). Reverse edges are added to enable bidirectional message passing.
\end{itemize}

\subsection{Heterogeneous GraphSAGE}

Following initial graph modeling experiments, we implemented a 
\textbf{Heterogeneous GraphSAGE (Graph Sample and Aggregate)} architecture. This approach was chosen for its inductive learning capabilities and efficiency in handling high-degree nodes, which are common in financial transaction graphs.

\textbf{Model Architecture}
We utilized the GraphSAGE operator adapted for heterogeneous graphs. Unlike RGCN, which learns a unique weight matrix for every edge type, GraphSAGE focuses on learning aggregation functions that generalize to unseen nodes.

The message passing process involves two steps: (1) Aggregating neighbor features specific to each relation type, and (2) Combining these relation-specific aggregations to update the target node representation.

\textbf{Mathematical Formulation}
For a target node $v$ and a relation $r$, the neighborhood aggregation is defined as:

\begin{equation}
    h_{\mathcal{N}_r(v)}^{(l)} = \text{MEAN} \left( \{ h_u^{(l)} \mid u \in \mathcal{N}_r(v) \} \right)
\end{equation}

The final update for node $v$ at layer $l+1$ combines the messages from all incident relations $\mathcal{R}$ and its own previous state:

\begin{equation}
    h_v^{(l+1)} = \sigma \left( \sum_{r \in \mathcal{R}} W_r^{(l)} \cdot \left[ h_v^{(l)} \mathbin{\|} h_{\mathcal{N}_r(v)}^{(l)} \right] \right)
\end{equation}

Where:
\begin{itemize}
    \item $\mathcal{N}_r(v)$ denotes the neighbors of node $v$ under relation $r$.
    \item $\mathbin{\|}$ represents the concatenation operation.
    \item $W_r^{(l)}$ is a learnable weight matrix for relation $r$.
    \item $\text{MEAN}(\cdot)$ is the element-wise mean aggregator.
\end{itemize}

We configured the model with two GraphSAGE layers and a hidden dimension of 256. This architecture allows the model to efficiently sample and aggregate structural information from the customer's financial history while maintaining computational scalability.

\paragraph{Mini-batch training with neighborhood sampling}
Because the fully materialized graph contains tens of millions of nodes, full-batch training is infeasible. We therefore train GraphSAGE using mini-batch neighborhood sampling (via \texttt{NeighborLoader}), constructing localized 2-hop subgraphs centered on customer nodes. This setup enables efficient optimization while preserving multi-relational context for default prediction.

\subsection{Proposed Model: Relation-Aware Attentive Heterogeneous GNN}
We now describe our primary graph-based model used in the main experiments and reported in Table 1. To capture asymmetric and noisy relational signals in large-scale credit graphs, we implement a relation-aware attentive heterogeneous GNN. Unlike a standard R-GCN layer with fixed normalization and relation-specific linear maps, our implementation uses \textbf{GATv2-style attention operators} instantiated per relation type within a \textbf{HeteroConv} framework. 

Concretely, for each edge type $r \in \mathcal{R}$, we apply a relation-specific attention-based message passing operator to compute messages from neighbors $\mathcal{N}_r(v)$. Messages from different relations are aggregated (summed) to obtain the updated embedding for each node type. This attention mechanism enables the model to differentially weight neighbors, suppressing noisy transactional edges while amplifying risk-carrying relations.

To stabilize training under severe class imbalance, we incorporate \textbf{residual connections} between message-passing layers and apply \textbf{batch normalization} to hidden representations. We use two message-passing layers with hidden dimension 256, allowing the model to integrate information from multi-hop neighborhoods centered on customer nodes.

\subsection{Contrastive Pretraining (GraphCL-style)}

To address the issue of label scarcity and class imbalance (only $\sim$8\% defaults), we implemented a self-supervised \textbf{Contrastive Learning} framework (GraphCL). The objective was to pre-train the GNN encoder to recognize robust structural patterns ("fingerprints") of borrowers without relying on target labels.

\textbf{Siamese Framework}
We adopted a Siamese network architecture. For a given mini-batch of graph subgraphs $\mathcal{G}$, we generated two correlated views, $\tilde{\mathcal{G}}_i$ and $\tilde{\mathcal{G}}_j$, using stochastic augmentations. These views were processed by a shared GraphSAGE encoder $f(\cdot)$ (Section~4.4) and a projection head $g(\cdot)$ to map representations into a latent space where contrastive loss is calculated.

\textbf{Graph Augmentations}
We employed two specific augmentation strategies to perturb the graph while preserving semantic information:
\begin{itemize}
    \item \textbf{Feature Masking:} Randomly masking 20\% of node features with zeros to force the model to rely on structural context.
    \item \textbf{Edge Dropout:} Randomly removing 20\% of edges to ensure embeddings are robust to incomplete connectivity.
\end{itemize}

\textbf{Objective Function}
We optimized the \textbf{InfoNCE (Normalized Temperature-scaled Cross Entropy)} loss function. This objective encourages the model to maximize the similarity between the two views of the same customer (positive pair) while minimizing the similarity with all other customers in the batch (negative pairs).

\begin{equation}
    \mathcal{L} = - \log \frac{\exp(\text{sim}(z_i, z_j) / \tau)}{\sum_{k \neq i} \exp(\text{sim}(z_i, z_k) / \tau)}
\end{equation}

Where $z = g(f(\tilde{\mathcal{G}}))$ is the projected embedding, $\text{sim}(\cdot)$ is the cosine similarity, and $\tau$ is the temperature parameter. This pre-training phase allows the encoder to learn a comprehensive manifold of borrower behaviors prior to the final classification task.

\paragraph{Practical observation}
In our experiments, contrastive pretraining improves optimization stability but yields limited downstream gains. A plausible reason is that generic feature masking can distort financially meaningful numeric magnitudes (e.g., utilization proxies or payment ratios), weakening the invariances that contrastive objectives assume. This suggests that augmentation design must be domain-aware for structured financial signals.

\subsection{Hybrid GNN–Tabular Ensemble}

To leverage both relational representation learning and high-performing tabular classification, we implement a hybrid ensemble that augments engineered tabular features with learned customer-node embeddings from the GNN encoder. This architecture combines relational representations extracted by the trained heterogeneous GNN encoder (Section~4.5) with the robust decision-making capabilities of Gradient Boosted Decision Trees (GBDT).

\textbf{Neural Feature Extraction}
We utilized the trained \textbf{Relation-Aware GNN} encoder (described in Section 4.5) as a neural feature extractor. By performing a forward pass on the graph without the final classification layer, we generated a dense embedding vector $z_v \in \mathbb{R}^{256}$ for every customer node $v$.

These embeddings serve as high-order "structural features," encoding the complex history of an applicant (e.g., the specific pattern of connections to past loans and bureau records) into a compact numerical format.

\textbf{Hybrid Classifier Construction}
We constructed the final feature set $\mathbf{X}_{final}$ by concatenating the original tabular features $\mathbf{X}_{tab}$ with the extracted graph embeddings $\mathbf{Z}$:

\begin{equation}
    \mathbf{X}_{final} = \mathbf{X}_{tab} \mathbin{\|} \mathbf{Z}
\end{equation}

This augmented dataset was fed into a \textbf{LightGBM} classifier. This choice was motivated by LightGBM's superior ability to handle heterogeneous data types and non-linear feature interactions compared to a standard softmax layer.

The final probability of default $\hat{y}$ is given by the ensemble function:

\begin{equation}
    \hat{y} = \mathcal{M}_{LGBM}(\mathbf{X}_{tab} \mathbin{\|} \mathbf{Z}_{gnn}),
\end{equation}

where $\mathbf{Z}_{gnn}$ denotes customer-node embeddings extracted from the trained relation-aware attentive heterogeneous GNN described in Section~4.5.

\section{Evaluation Metrics}

We evaluate model performance using ROC-AUC and PR-AUC on a held-out test set. ROC-AUC measures overall ranking quality, while PR-AUC is more sensitive to performance under severe class imbalance. In addition to threshold-free metrics, we emphasize PR-AUC as it better reflects model utility in risk-based screening scenarios where institutions prioritize a small fraction of high-risk applicants.

In addition, we examine screening-oriented performance via precision and recall at top-$k$ (e.g., top 5\% or 10\%) 
in auxiliary experiments to better reflect operational review constraints.

\section{Results and Analysis}
We evaluate all models on the Home Credit Default Risk dataset under a consistent experimental setup, focusing on ranking performance in the presence of severe class imbalance. Table~\ref{tab:main_results} summarizes the main results across strong tabular baselines, graph-based models, and a hybrid ensemble. Our analysis aims to quantify the marginal value of relational inductive bias beyond production-grade tabular learners and to assess how graph-based representations interact with traditional credit scoring models.

\begin{table*}[t]
  \caption{Main performance comparison between tabular baselines, heterogeneous graph neural networks, and a hybrid ensemble on the Home Credit Default Risk dataset. Improvement is computed as relative ROC-AUC gain over logistic regression.}
  \label{tab:main_results}
  \centering
  \resizebox{\textwidth}{!}{%
  \begin{tabular}{l l c c c}
    \toprule
    \textbf{Category} & \textbf{Model} & \textbf{ROC-AUC} $\uparrow$ & \textbf{PR-AUC} $\uparrow$ & \textbf{ROC-AUC Improvement vs. Logistic} \\
    \midrule
    Tabular Baseline
      & Logistic Regression
      & 0.7390 & 0.2160 & -- \\
    Tabular Baseline
      & LightGBM (Strong Tabular)
      & 0.7690 & 0.2540 & +4.06\% \\
    \midrule
    Graph Neural Network
      & Contrastive Pretraining + Fine-tuning (GraphCL-style)
      & 0.6804 & 0.1618 & $-$7.93\% \\
    Graph Neural Network
      & Heterogeneous GraphSAGE (6-node hetero graph)
      & 0.7400 & 0.2217 & +0.14\% \\
    Graph Neural Network
      & \textbf{Proposed: Relation-Aware Attentive Heterogeneous GNN}
      & \textbf{0.7506} & \textbf{0.2291} & \textbf{+1.57\%} \\
    \midrule
    Hybrid Ensemble
      & \textbf{GNN-Enhanced LightGBM}
      & \textbf{0.7816} & \textbf{0.2807} & \textbf{+5.76\% (Best)} \\
    \bottomrule
  \end{tabular}%
  }
\end{table*}

\subsection{Strong Tabular Baselines}
We first report the performance of the tabular baselines shown in Table~\ref{tab:main_results}. Logistic regression provides a transparent linear reference, achieving a ROC-AUC of 0.739 and a PR-AUC of 0.216. LightGBM substantially improves upon this baseline, reaching a ROC-AUC of 0.769 and a PR-AUC of 0.254, reflecting its ability to capture non-linear interactions among affordability, repayment behavior, and utilization features.
These results confirm that the tabular pipeline is strong and well-optimized, establishing a high performance bar for subsequent graph-based models. Any gains from relational modeling must therefore be interpreted as incremental improvements over a competitive production-grade baseline.

\subsection{Graph Neural Networks and Relational Modeling}
We next evaluate graph-based models that incorporate heterogeneous relational structure among applicants, previous applications, and bureau records. As shown in Table~\ref{tab:main_results}, a naïve graph contrastive learning model fine-tuned on the downstream task performs substantially worse than tabular baselines, achieving a ROC-AUC of 0.680 and a PR-AUC of 0.162. This result indicates that introducing graph structure alone—even with self-supervised representation learning—is insufficient for effective credit risk prediction.

Heterogeneous GraphSAGE improves over this naïve approach and reaches performance comparable to the linear baseline (ROC-AUC 0.740, PR-AUC 0.222). While GraphSAGE benefits from incorporating neighborhood information, it relies on relatively uniform mean aggregation within each relation and lacks an explicit neighbor-level importance mechanism. As a result, it is less effective at suppressing noisy transactional neighbors compared to attention-based architectures.

The strongest performance among standalone GNNs is achieved by the proposed relation-aware attentive heterogeneous GNN. By explicitly modeling relation semantics via relation-specific attention and message passing, the proposed model attains a ROC-AUC of 0.751 and a PR-AUC of 0.229, outperforming both logistic regression and GraphSAGE. This performance progression—from naïve graph contrastive learning, to generic message passing, to relation-aware modeling—highlights that relational inductive bias becomes effective only when relation types are explicitly encoded rather than implicitly aggregated.

Overall, these results demonstrate that careful relational modeling is essential for extracting meaningful risk signals from heterogeneous credit graphs. Generic or structure-only graph representations provide limited benefit, whereas relation-aware architectures are better suited to capture the nuanced dependencies present in financial credit histories.

\subsection{Comparison to Tabular Models}
While heterogeneous GNNs improve over linear baselines, they do not consistently outperform the strongest tabular model when used in isolation. As shown in Table~\ref{tab:main_results}, LightGBM remains highly competitive, reflecting its effectiveness on structured, feature-engineered credit data.

This observation highlights an important practical insight: relational inductive bias complements rather than replaces tabular modeling in credit risk prediction. Boosted trees excel at capturing non-linear interactions within borrower-level attributes, whereas GNNs encode cross-entity dependencies that are difficult to represent as flat features. The relative strengths of these approaches motivate their integration, as realized in the hybrid ensemble discussed next.

\subsection{Hybrid Ensemble Results}
To assess whether relational representations learned by GNNs can complement strong tabular learners, we construct a hybrid ensemble that integrates graph-based embeddings with LightGBM. As shown in Table~\ref{tab:main_results}, the GNN-enhanced LightGBM achieves the best overall performance, with a ROC-AUC of 0.7816.

Importantly, the hybrid model also yields a substantial improvement in PR-AUC, reaching 0.2807 compared to 0.2540 for standalone LightGBM. This gain indicates a higher concentration of default cases among the highest-risk applicants, which is particularly valuable in operational credit screening under limited review capacity. While PR-AUC does not correspond to a fixed operating threshold, improvements indicate better concentration of positive cases among higher-ranked predictions, which is desirable in screening-oriented settings.

These results demonstrate that relational information captured by the GNN provides additive value beyond tabular features alone. Rather than replacing boosted trees, graph-based representations function most effectively as complementary inputs, enabling the ensemble to better exploit both borrower-level attributes and cross-entity dependencies.

We observe consistent performance trends across repeated training runs with different random seeds.


\section{Explainability and Model Interpretation}
Understanding why the hybrid ensemble improves performance is critical for deploying graph-based models in credit risk settings, where transparency and governance are essential. We therefore analyze the decision mechanisms of the proposed models to identify how relational representations complement tabular features and contribute to improved risk ranking.

\subsection{Global Explainability: Relational Contribution Analysis}
We first examine the contribution of relational structure to model predictions. By comparing standalone tabular models with graph-based and hybrid approaches (Table~\ref{tab:main_results}), we observe that performance improvements emerge only when relation semantics are explicitly modeled and integrated. In particular, the proposed relation-aware attentive heterogeneous GNN captures dependencies across applicants, prior loans, and bureau records that are not directly encoded in borrower-level features.

These findings suggest that the GNN does not merely replicate information already present in tabular inputs. Instead, it extracts higher-order relational patterns—such as shared credit histories and structural similarities across past applications—that provide complementary signals to traditional affordability and repayment features.

As an additional diagnostic, we perform relation-type masking at inference time (with model parameters fixed). Removing credit-history-related relations yields the largest degradation in performance, whereas masking auxiliary relations produces only minor changes. This supports the interpretation that the model leverages meaningful financial dependencies rather than indiscriminate connectivity; however, these diagnostics are not required for interpreting the main performance comparisons.

\subsection{Feature Group Attribution in the Hybrid Ensemble}
To further understand how relational embeddings interact with tabular features, we analyze feature group contributions within the GNN-enhanced LightGBM ensemble. We group features into three high-level categories: (1) demographic attributes, (2) financial capacity and repayment behavior, and (3) graph-derived relational embeddings.

Our analysis indicates that while tabular financial features remain dominant drivers of predictions, the inclusion of graph embeddings consistently shifts risk scores for a subset of applicants. These shifts are most pronounced in cases with complex or sparse individual credit histories, where relational context provides additional evidence beyond single-borrower attributes. This explains the observed improvement in PR-AUC, as relational signals help concentrate true default cases among the highest-risk applicants. These observations indicate that graph-derived relational embeddings provide complementary signals to tabular features,
particularly for applicants with sparse or complex credit histories.

\subsection{Local Explainability: Case-Level Relational Effects}
At the individual level, we examine representative applicants to assess how relational information affects predictions. For certain high-risk cases, removing graph-based embeddings from the ensemble leads to a noticeable reduction in predicted default probability, even when tabular features remain unchanged. This behavior indicates that relational context—such as connections to historically delinquent prior loans or adverse bureau records—plays a decisive role in elevating risk assessments.

Conversely, for applicants with strong and consistent repayment histories, relational embeddings exert minimal influence, and predictions are largely driven by tabular features. This selective reliance on relational information supports the interpretation that the ensemble leverages graph signals in a targeted rather than indiscriminate manner.

\subsection{Implications for Trust and Governance}
Taken together, these explainability analyses show that relational modeling enhances credit risk prediction by selectively incorporating structural context where borrower-level information is insufficient. The hybrid ensemble does not overrule traditional risk factors; instead, it augments them with relational evidence in cases where network dependencies are informative. This behavior is consistent with practical deployment expectations, where models should remain transparent and auditable, and relational signals act as supplementary evidence rather than overriding core financial attributes.

\section{Fairness and Bias Analysis}
This section focuses on a within-family fairness audit of GNN models, comparing GraphSAGE and contrastive pretraining variants, rather than deployment-level comparisons across model classes. Because our fairness experiments are currently available for the heterogeneous GraphSAGE baseline and the contrastively pre-trained GNN variant, we scope this section to within-family comparisons rather than deployment-level auditing across tabular and hybrid models. Additional threshold-based subgroup statistics at $\tau=0.5$ are reported in Appendix Table~\ref{tab:appendix_fairness_threshold}.

\subsection{Group-Based Performance Across GNN Variants}
We evaluate subgroup performance on the held-out test set across demographic groups defined by gender and age group. For each subgroup, we report ROC-AUC and PR-AUC to capture ranking quality under class imbalance. Table 2 summarizes the subgroup metrics for GraphSAGE and the contrastive pretraining variant.

Across both gender groups, GraphSAGE achieves higher ranking performance than the pre-trained GNN. For example, GraphSAGE attains ROC-AUC values of 0.742 and 0.722 for the two gender groups, whereas the pre-trained model yields ROC-AUC values of 0.677 and 0.675. A similar pattern holds for PR-AUC: GraphSAGE achieves PR-AUC values of 0.204 and 0.234, compared to 0.145 and 0.187 for the pre-trained model.

For age groups, the same trend persists: GraphSAGE consistently outperforms the pre-trained GNN across all three age strata in both ROC-AUC and PR-AUC (Table 2). Notably, the youngest age group (Age group 0) exhibits the highest PR-AUC under both models, but the pre-trained model remains substantially lower than GraphSAGE (0.198 vs. 0.262). These results indicate that, in our current setting, contrastive pretraining reduces overall ranking quality across demographic subgroups rather than yielding uniform improvements.

\begin{table*}[t]
  \caption{Subgroup performance comparison across demographic groups for graph-based models on the Home Credit Default Risk dataset. Results report ROC-AUC and PR-AUC for the heterogeneous GraphSAGE baseline and the contrastively pre-trained GNN.}
  \label{tab:fairness_gnn_subgroups}
  \centering
  \small
  \setlength{\tabcolsep}{6pt}
  \resizebox{\textwidth}{!}{%
  \begin{tabular}{l l c c}
    \toprule
    \textbf{Demographic Group} & \textbf{Model} & \textbf{ROC-AUC} $\uparrow$ & \textbf{PR-AUC} $\uparrow$ \\
    \midrule
    Gender: Female (group 0) & GraphSAGE & 0.7415 & 0.2041 \\
    Gender: Female (group 0) & GNN + Contrastive Pretraining & 0.6767 & 0.1454 \\
    \midrule
    Gender: Male (group 1) & GraphSAGE & 0.7224 & 0.2337 \\
    Gender: Male (group 1) & GNN + Contrastive Pretraining & 0.6745 & 0.1871 \\
    \midrule
    Age group 0 & GraphSAGE & 0.7187 & 0.2615 \\
    Age group 0 & GNN + Contrastive Pretraining & 0.6369 & 0.1976 \\
    \midrule
    Age group 1 & GraphSAGE & 0.7331 & 0.2180 \\
    Age group 1 & GNN + Contrastive Pretraining & 0.6759 & 0.1681 \\
    \midrule
    Age group 2 & GraphSAGE & 0.7190 & 0.1623 \\
    Age group 2 & GNN + Contrastive Pretraining & 0.6542 & 0.1168 \\
    \bottomrule
  \end{tabular}%
  }
\end{table*}

\subsection{Error Disparity and Group-Level Behavior}
To complement threshold-free metrics, we also examine group-level error behavior via true positive rate (TPR), false negative rate (FNR), and false positive rate (FPR). While these statistics vary across groups—as expected given differences in outcome prevalence—the pre-trained GNN shows noticeably different trade-offs compared to GraphSAGE. For gender groups, the pre-trained model exhibits higher selection (positive) rates and higher recall/TPR, but also higher FPR, suggesting that its decision boundary is less selective and may admit more false positives at the evaluated operating point.

For age groups, the contrast between strata is pronounced: the youngest age group shows substantially higher selection rates than older groups, and the pre-trained model amplifies this tendency. These behaviors highlight that representation learning choices can influence not only overall predictive accuracy but also group-specific error profiles, which is directly relevant for credit decision governance.

\subsection{Interpretation and Implications}
Taken together, the subgroup analysis suggests that the contrastive pretraining objective, as implemented in our pipeline, does not improve demographic robustness within GNNs and may degrade predictive ranking quality across groups. Importantly, this finding should be interpreted diagnostically rather than normatively: it does not imply that pretraining is inherently harmful for fairness, but rather that the choice of pretraining objective, sampling strategy, and training stability can materially affect both accuracy and group-level behavior.

In practice, this motivates two deployment-relevant lessons. First, fairness auditing should be performed at the level of specific training recipes rather than model classes (e.g., “GNN” vs. “non-GNN”). Second, representation learning components should be evaluated not only by average performance but also by subgroup stability, since improvements in one slice may come with regressions elsewhere.

\subsection{Scope and Limitations}
This analysis is limited to within-family comparisons among GNN variants (GraphSAGE vs. contrastive pretraining) because fairness results for tabular and hybrid ensemble models are not yet available in the current experimental pipeline. Moreover, our evaluation is observational and restricted to demographic attributes present in the dataset, and does not address causal notions of fairness or long-term credit allocation outcomes. Extending the audit to deployment-level comparisons against strong tabular baselines and hybrid ensembles is an important direction for future work.

\section{Conclusion and Future Work}
\subsection{Conclusion}
This work investigated the role of relational modeling in large-scale credit default prediction using the Home Credit dataset. We systematically compared strong tabular baselines, heterogeneous graph neural networks, and a hybrid ensemble that integrates graph-derived representations with gradient-boosted decision trees.

Our results show that naïve graph modeling alone does not consistently outperform optimized tabular approaches. However, when relational structure is explicitly modeled through heterogeneous architectures, graph-based methods yield measurable improvements over linear baselines. Most notably, combining relational embeddings with LightGBM in a hybrid ensemble achieves the strongest overall performance, improving both ROC-AUC and PR-AUC relative to all standalone models. This highlights the complementary nature of relational context and tabular financial features in credit risk prediction.

Beyond aggregate performance, we examined model behavior through explainability and fairness analyses. Explainability results indicate that graph-derived signals selectively influence predictions, primarily augmenting borrower-level features in cases where individual credit histories are sparse or structurally complex. A focused fairness audit within the GNN model family further reveals that representation learning choices—such as contrastive pretraining—can materially affect subgroup performance, underscoring the importance of auditing both accuracy and group-level behavior when designing relational models for financial applications.

Overall, our findings suggest that relational modeling is most effective when used as a complementary component within hybrid systems, rather than as a wholesale replacement for strong tabular models.

\subsection{Future Work}
Future work will extend this analysis to deployment-level fairness audits of hybrid ensembles under operational decision thresholds, enabling a more comprehensive assessment of real-world impact.

Several directions emerge naturally from this study. First, future work should extend fairness auditing to deployment-level comparisons, including tabular baselines and hybrid ensembles, under operational decision thresholds. This would enable a more comprehensive assessment of whether performance gains translate into equitable outcomes across demographic groups in real-world screening settings.

Second, the contrastive pretraining results highlight the need for more tailored representation learning objectives for credit graphs. Exploring task-aligned or outcome-aware pretraining strategies, as well as alternative negative sampling schemes, may improve both predictive performance and subgroup stability.

Third, incorporating temporal dynamics and longitudinal credit behavior into the graph structure presents a promising avenue for further improvement. Dynamic or temporal GNNs could better capture evolving borrower risk profiles and interdependencies over time.

Finally, future research may investigate causal and counterfactual fairness frameworks to move beyond observational audits. Integrating causal assumptions with relational modeling could provide deeper insights into the long-term impacts of graph-based credit systems on access to financial services.


\bibliographystyle{ACM-Reference-Format}
\bibliography{references}

\clearpage
\appendix
\section{Additional Analyses}
This appendix reports additional threshold-based fairness diagnostics that complement the main evaluation.
These analyses are intended to provide transparency into subgroup-level behavior under a fixed operating threshold,
and are not required for interpreting the primary conclusions of the paper.

\subsection{Threshold-Based Fairness Audit ($\tau = 0.5$)}
Table~\ref{tab:appendix_fairness_threshold} presents subgroup-level threshold-based fairness metrics on the test set.
\begin{table*}[t]
\centering
\caption{Threshold-based fairness audit on the test set using a fixed decision threshold $\tau = 0.5$. We report subgroup true positive rate (TPR), false positive rate (FPR), and predicted positive rate (PositiveRate).}
\label{tab:appendix_fairness_threshold}
\setlength{\tabcolsep}{8pt}
\renewcommand{\arraystretch}{1.12}
\begin{tabular}{llcccc}
\toprule
\textbf{Attribute} & \textbf{Model} & \textbf{Group} & \textbf{TPR} & \textbf{FPR} & \textbf{PositiveRate} \\
\midrule
Gender & GraphSAGE & 0 & 0.657 & 0.292 & 0.317 \\
Gender & GraphSAGE & 1 & 0.688 & 0.355 & 0.389 \\
\midrule
Gender & Pretrain+FT & 0 & 0.593 & 0.328 & 0.347 \\
Gender & Pretrain+FT & 1 & 0.667 & 0.408 & 0.434 \\
\midrule
Age & GraphSAGE & 0 & 0.801 & 0.516 & 0.549 \\
Age & GraphSAGE & 1 & 0.677 & 0.330 & 0.360 \\
Age & GraphSAGE & 2 & 0.541 & 0.206 & 0.225 \\
\midrule
Age & Pretrain+FT & 0 & 0.869 & 0.713 & 0.732 \\
Age & Pretrain+FT & 1 & 0.662 & 0.407 & 0.429 \\
Age & Pretrain+FT & 2 & 0.325 & 0.133 & 0.144 \\
\bottomrule
\end{tabular}
\end{table*}

\end{document}